\documentclass[runningheads]{llncs}
\usepackage[T1]{fontenc}
\usepackage{graphicx}
\begin{document}
\title{Hateful Messages: A Conversational Data Set of Hate Speech produced by Adolescents on Discord\thanks{This research was supported by the Citizens, Equality,
Rights and Values (CERV) Programme under Grand Agreement No. 101049342.}}

\titlerunning{Hateful Messages}
%
\author{Jan Fillies\inst{1,3} \and
Silvio  Peikert\inst{2}\and Adrian  Paschke \inst{1,2,3}}

\authorrunning{J. Fillies et al.}
%

\institute{Institut für Angewandte Informatik\\\and
Fraunhofer-Institut für Offene Kommunikationssysteme FOKUS\\\and
Freie Universität Berlin
}

%
\maketitle              
\begin{abstract}
With the rise of social media, a rise of hateful content can be observed. Even though the understanding and definitions of hate speech varies, platforms, communities, and legislature all acknowledge the problem. Therefore, adolescents are a new and active group of social media users. The majority of adolescents experience or witness online hate speech. Research in the field of automated hate speech classification has been on the rise and focuses on aspects such as bias, generalizability, and performance. To increase generalizability and performance, it is important to understand biases within the data. This research addresses the bias of youth language within hate speech classification and contributes by providing a modern and anonymized hate speech youth language data set consisting of 88.395 annotated chat messages. The data set consists of publicly available online messages from the chat platform Discord. ~6,42\% of the messages were classified by a self-developed annotation schema as hate speech. For 35.553 messages, the user profiles provided age annotations setting the average author age to under 20 years old.

\keywords{Hate Speech  \and Youth Language \and Bias \and Data Set\and NLP.}
\end{abstract}
%
%
\section{Introduction}
\label{intro}
Research shows that there are differences within the language used by age groups online \cite{Schwartz2013}. Most teenagers within the United States, use Social media \cite{Thapa_Subedi_2018}. In the span of January 2020 till March 2020 Facebook removed 9.6 million posts containing hate speech\footnote[1]{https://www.forbes.com/sites/niallmccarthy/2020/05/13/facebook-
removes-record-number-of-hate-speech-posts-infographic/?sh=20c0ef983035}. As of today, it is clear that social media is used often and frequently by adolescents. Hate speech and its algorithmic detection has had an increasing interest in social media platforms such as e.g. Facebook\footnote[2]{https://ai.facebook.com/blog/how-facebook-uses-super-efficient-ai-models-to-detect-hate-speech}. This development is especially supported by the harmful effects hate speech has on it recipients \cite{Saha2019}. 

Based on the research that identifies a difference in language, time, and topic in conversations between adolescents and adults \cite{Schwartz2013}, it is necessary to build a database of youth language to explore the impact the language has upon algorithmic hate speech detection. This research lays the groundwork to close the gap by introducing an annotated hate speech data set focusing on youth language. The data set was collected in a real-world environment in 03.2021 – 06.2022. It provides the scientific community a modern corpus that can be used to evaluate the bias in existing classification algorithms for hate speech and further train domain specific algorithms to the setting of hate speech within the online chat conversations of adolescents. This modern real world data set overcomes the status quo of identifying hate speech connected to geolocation and introduces the view that the hate speech is also unique to international group conversations on the internet.


\section{Related Literature}
\label{related}

Annotated data sets in the field of hate speech detection are available (e.g. \cite{Hosseinmardi2015,Gibert2018}. There are fewer multilingual data sets with fitting annotations available \cite{Chung2019}. Hate speech data sets have many annotation schemes \cite{Chung2019}, from binary to multi-class hierarchies. Other universal annotation schemes exist \cite{Bartalesi2006} and are deployed in hate speech annotation or similar contexts, such as cyberbullying \cite{Sprugnoli2018}. It is difficult to obtain hate speech data sets, and hate speech information within adolescents. Research focusing on cyberbullying in pre-teens can be found in the instance of Sprugnoli et al. \cite{Sprugnoli2018}.  Sprugnoli et al. \cite{Sprugnoli2018} created a data set containing annotated hate speech chat conversations between Italian high school students. The data set was created in an experimental setting to foster a safe environment, moderated by the researcher. In 2019, Menini et al. \cite{Menini2019} presented a monitoring system for cyberbullying. They identified a network of multiple high schools, their students and their friends in the United Kingdom’s Instagram community. In 2020, Wijesiriwardene et al. \cite{Wijesiriwardene2020} published a multimodal data set containing Tweets labeled for toxic social media interactions. The data set was created focusing on American high school students. In 2011 Bayzick \cite{Bayzick2011} created a data set consisting of messages from MySpace.com. They organized the messages into groups of ten and annotated the messages, some of which contained cyberbullying. The data set includes self-provided information about the age of the author. Dadvar et al. \cite{Dadvar2013} showed that user context including attributes such as age, gender, and cyberbullying history of the user improves the detection of cyberbullying. Chen et al. \cite{Chen2012} analysis the personal writing style and other user specific attributes to identify the potential of the user spreading hate speech. 

Natural language processing is required to be algorithmically fair and fitted to many social groups \cite{Blodgett2017}. Classification algorithms can be biased towards many minority groups of people. For example, bias by gender \cite{Kurita2019} or race \cite{Kennedy2020}. Even though age is a known source for bias in data \cite{Hovy2021}, it is not widely analyzed in pretrained networks. To counter these biases, there are different approaches. Some focus on single domains or tasks via fine-tuning using new data \cite{Park2018}. 

As shown, there are numerous publicly available hate speech data sets, some are addressing the adolescent audience and are annotated for cyberbullying or hate speech. But there are three missing fields. Firstly, our research focuses on online conversations, not comments under posts. Secondly, the introduced dataset is drawn out of a real-world setting and not created in an artificial experimental setting. Thirdly, this data focuses on an international online English speaking community, not a regional community. Lastly the aspect of time needs to be considered, it is necessary to collect and analyze the data sets from a recent time frame, considering the shifts in topic and language. 

Within the last five years, no real-world hate speech data set containing online conversations of adolescents could be found. 

\section{Hate Speech Data Set}
\label{data}

\textbf{Methodology}
Vidgen and Derczynski \cite{bertie2021} recommend addressing the following points when creating an abusive content data set. “Purpose” the purpose of this data set is to build a base for validation and improving hate speech detection within youth language, further explained in section \ref{intro}. “Explore new source”, no hate speech Discord data set has been discovered so far. “Clear taxonomy” the used taxonomy is based on Paasch-Colberg et al. \cite{PaaschColberg2021} and in detail described later. “Develop guidelines iteratively with your annotators” this has been done and is described in section Annotations Procedure. “Data Statement” a data statement is provided using the format suggested by Gebru et al. \cite{Gebru2021}.

\textbf{Data Identification,}
Discord is a chat platform that provides spaces for communication between users. These servers are publicly available and if configured can be joined by anyone interested. There are public lists available for existing chat servers, filterable by language, name, and topics. The research project pre-selected a list of servers by identifying problematic sounding groups names, and then selected the annotated chat room by evaluating them based on the following five criteria. Firstly, conversation language in English. Secondly, high appearance of general derogatory terms through a simple key word search. Thirdly, amount of active users. Fourthly, amount of messages send in the group chat. And lastly, available information on the age of the users. The chosen chat room fulfills these criteria and was exported for the purpose of furthering research within the topic.

\textbf{Annotation Guidelines,}
\label{anno}
The annotation guidelines were developed iteratively with and by the annotators, ensuring a high understanding of the process and definitions. A common definition of hate speech was established. Hereby, a statement is viewed as hate speech if it is directed towards a group or an individual of a group with the characteristic of excluding and stigmatizing. A statement is further considered hate speech if it is hostile, implies the desire to harm or incite violence. Based on Paasch-Colberg et al. \cite{PaaschColberg2021} a new annotation schema was defined including descriptions and examples. All nine categories of the schema are explained in the following.  
\newcommand\Tstrut{\rule{0pt}{2.0ex}}         
\newcommand\Bstrut{\rule[-1.2ex]{0pt}{0pt}}   
\newcommand\TBstrut{\Tstrut\Bstrut}           

\begin{table}
\scriptsize
\centering
\caption{Table of annotated classes.}\label{tab0}
\begin{tabular}{|l|l|l|}
\hline
Label & Definition & Example \\
\hline
No Hate &  \parbox{5cm}{\Tstrut Positive and neutral conversations, but also criticism, rejection and disliking\TBstrut}  & \parbox{4cm}{"I don't like the new chairs"}\\\hline
Negative Stereotyping &   \parbox{5cm}{\Tstrut Generalizations in which hurtful intent is a central motivator\TBstrut}  & \parbox{4cm}{"All blondes are stupid"}\\\hline
Dehumanization & \parbox{5cm}{\Tstrut Non-human character traits are attributed to humans, people, or groups. They are paired with elements not belonging to the human species\TBstrut} & \parbox{4cm}{"Asian rats"} \\\hline
Violence and Killing & \parbox{5cm}{\Tstrut Endorse, glorify, or fantasize about violence or killing, the explicit call to violence and murder\TBstrut} & \parbox{4cm}{"The only thing that [...] helps is pure violence"}\\\hline
Equation & \parbox{5cm}{\Tstrut Associates’ people and groups of people with negative characteristics\TBstrut} & \parbox{4cm}{"Poor = Africa"} \\\hline
Norm. of Exi. Dis. & \parbox{5cm}{\Tstrut Existing discrimination is downplayed and or manifested\TBstrut} & \parbox{4cm}{"No wonder blacks are treated this way."} \\\hline
Disguise as Irony & \parbox{5cm}{\Tstrut Disguises negative statements as irony and downplays them as humor\TBstrut} & \parbox{4cm}{"In my next life I'll be a social welfare recipient, there I can chill"}\\\hline
Harmful Slander & \parbox{5cm}{\Tstrut All other forms of insults and hurtful statements that cannot be classified into the previous labels\TBstrut} & \parbox{4cm}{"I don't know any "normal" Jew"} \\\hline
Skip & \parbox{5cm}{\Tstrut Comments that cannot be understood and not assigned to labels above, due to a linguistic or symbolic barrier\TBstrut} & \parbox{4cm}{Examples of this are emojis or Asian characters} \\
\hline
\end{tabular}
\end{table}

\textbf{Annotations Procedure,}
\label{process}Five annotators have annotated the data set. The team of annotators consisted of Bachelor and Master computer science students, and the average age of the members was 29 years. The ages varied from 21 to 58 years. The team consisted of two female and three male annotators. For four out of five members, the ethnic background and mother tongue was German. One annotator's mother tongue and ethnic background was Albanian. One group member brought domain-specific knowledge through a degree as a translator. 

The data set was divided into equal parts so that simultaneous annotation was possible. An annotation tool was used. Messages that were uncertain or not clear for the annotator were jointly annotated in the peer review process. 

\textbf{Data Statement,}
The statement is provided in table \ref{tab01} and based on Gebru et al. \cite{Gebru2021}. The classes "RECORDING QUALITY, "OTHER", and "PROVENANCE APPENDIX" were not available or applicable for the data set.
\label{datastate}
\begin{table}
\scriptsize
\centering
\caption{The Data Statement}\label{tab01}
\begin{tabular}{|l|l|}
\hline
Characteristic & Description \\
\hline
\parbox{3cm}{CURATION \\RATIONALE} &  \parbox{9cm}{\Tstrut The data set consists of Discord chat massages. The chat room was selected due to the high level of hate speech, the age of the authors and the platform's young user base\TBstrut}  \\\hline
\parbox{3cm}{LANGUAGE\\VARIETY} &   \parbox{9cm}{\Tstrut The messages are online, written in English, and provided by a multinational setting, predominantly Europe, United Kingdom and America. It is partly youth language\TBstrut}\\\hline
\parbox{3cm}{SPEAKER \\DEMOGRAPHIC} & \parbox{9cm}{\Tstrut There are 249 unique author IDs. In the data set and, the average age of the user who provided information was under 20. Out of the users who provided information, 19 are female and 22 male. Out of the users who provided information, 15 were from the UK, 18 from the USA, 17 from Europe, and 13 from other countries. Disordered speech is present.\TBstrut} \\\hline
\parbox{3cm}{ANNOTATOR \\DEMOGRAPHIC} & \parbox{9cm}{\Tstrut Five annotators were used. They are full time students with an age range between 21-58, average age 29. The group consisted of three males, two females, four native German speakers, one native Albanian speaker. Four members were German natives and one member was an native Albanian. Lastly, one member holds a degree as a translator.\TBstrut} \\\hline
\parbox{3cm}{SPEECH \\SITUATION} & \parbox{9cm}{\Tstrut The data set was collected between 26.03.21 and 15.06.2022. It consists of written unscripted asynchronous messages. And the intended audience were the other participants in the chat.\TBstrut}\\\hline
\parbox{3cm}{TEXT \\CHARACTERISTICS} & \parbox{9cm}{\Tstrut It is an everyday conversational setting which guarantees its members free speech\TBstrut} \\\hline

\end{tabular}
\end{table}

\section{Data Evaluation}
The evaluation is oriented on the data evaluation performed by de Gilbert et al. \cite{gilbert2018}.
\label{eval} This research presents a novel English hate speech data set containing online conversations of adolescents on Discord. All messages have time stamps and author id's attached. The users are from differed countries, mainly USA, EU, GB. The data set consists of 88.395 messages. Out of these, 35.553 have an age annotation available and 52.895 do not. Table~\ref{tab1} shows the distribution of messages over all nine annotated categories. It is visible that the classes are not balanced, with most classes having less than 1000 messages assigned and the non hate speech class dominates with over 87\% of the data set. The whole data set contains 6,41\% hateful messages and the age annotated subset contains 5,07\% hateful messages. There are 9 members in the age group 14-17, 19 members in the age group 18-25, and 4 members in the age group 26+.
\begin{table}
\scriptsize
\centering
\caption{Table with distribution of labels.}\label{tab1}
\begin{tabular}{|l|l|l|l|l|}
\hline
Label & Total Messag. & \% Total Messag. & Age Messag. & \% Age Messag.\\
\hline
No Hate &  77.034 & 87,15 & 31.037 & 87,30\\
Negative Stereotyping &  768 & 0,9 & 232 & 0,7\\
Dehumanization & 499 & 0,6 & 165 & 0,5\\
Violence and Killing & 651 & 0,7 & 205 & 0,6\\
Equation & 124 & 0,1 & 27 & 0,08\\
Norm. of Exi. Dis. & 145 & 0,2 & 40 & 0,1\\
Disguise as Irony & 181 & 0,2 & 60 & 0,2 \\
Harmful Slander & 3.303 & 3,7 & 1.075 & 3,0\\
Skip & 5.689 & 6,4 & 2.712 & 7,6\\
Total & 88.395 & 100 & 35.553 & 100\\
\hline
\end{tabular}
\end{table}

The distribution of the comments in relation to the 249 registered users shows directly that 90\% of all messages were written by 30 users. On average, a user posts 2662 messages to the chat room, and 90\% of the hate speech was produced by 85 users with an average user posting 60 hateful messages. It was discovered, that one highly active user is making up 33.372 messages accounting for 2488 or 43,87\% of the hateful messages. This user is not classified as a chatbot and did not provide data regarding their age, therefore is not influencing the age annotated data sub set. In the age annotated sub set, 90\% of the messages are attributed to 10 users with 35 users providing 90\% of the hateful messages in the sub set, sending an average of 54 hateful messages per user.

Based on de Gilbert et al. \cite{gilbert2018} a hate score (HS) for each word (w) has been calculated as a simple way to create insight into the context in which a word appears. For this, all hateful classes (hate) have been combined to one. A Pointwise Mutual Information (PMI) score has been calculated between each word, the hateful class and the non-hateful (nohate) class. Then, the hate score of each word was calculated by subtracting the non-hateful PMI from the hateful PMI. 
\begin{equation}
\small
HS(w) = PMI(w, hate) - PMI(w,nohate)
\end{equation}
The table \ref{tab2} displays the five words with the highest  and lowest correlation to the class hateful. The hateful terms were modified to reduce their impact on the reader. The hateful words are strong, well known, slurs and deformations. The least correlated are surprisingly lead by an emoticon followed by general words including the word "fair". In these ten words, a youthful character can be identified, for example, by the extensive emoticon use and the usage of hateful modern abbreviation as "kys".
\begin{table}
\scriptsize
\centering
\caption{Table with the most positive (hateful) \&  most negative (least hateful) HS.}\label{tab2}
\begin{tabular}{|l|l|l|l|}
\hline
Hateful Word & HS & Non-Hateful Word & HS \\
\hline
ni**er &  11,0493 & \includegraphics[scale=0.5]{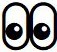} & -4.4188 \\
b*tch &  9,6315 & fair & -4.0526\\
fa*got & 9,2531 & \includegraphics[scale=0.5]{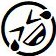} & -3.6197\\
h*e & 9,1535 & plus & -3.4836\\
kys & 7,4165 & huh & -3.4132\\
\hline
\end{tabular}
\end{table}
\section{Discussion}
\label{dis}
It is important to understand that online chat rooms, like the one evaluated, are an ecosystem, meaning the users influence each other in language and topic. Therefore, this youth language corpus might be fundamentally different from other youth language corpora. The approach used in this paper is still an important contribution to the world of youth language data sets due to the use of modern language with the provided self-identification, putting the general discussion in an age range under 20.
There is no way to guarantee that the given age ranges are truthful. This limitation cannot be easily circumvented in a non-experimental setting if a real-world data set is wanted while data protection is being guaranteed. 
During the creation of the data set following the official recommendation, the classes of the annotation schema were developed in communication with the annotators. This led to an arguable improvable annotation schema. The work is open to change and updates in the class definitions or reannotation. 
The data set is heavily unbalance in sense of authorship of the messages and the labeled hate speech classes. This is due to the real-life character of the data set and is a common problem in the field of hate speech.
It is important to start collecting and publishing subdomain data sets to understand the difference and uniqueness of languages in these groups and identify best performing hate speech classification algorithms.

\section{Conclusion and Future Work}
\label{conc} This paper collected and annotated a youth language data set containing 88.395 online chat messages. 
Of the 249 unique users, 31 provided information about their age, averaging to under 20 years. The data set is labeled into 9 classes in the field of hate speech. A data analysis has been conducted and influential terms for the "Hate" and "No Hate" classes have been established. A data statement is provided. The data set is available for scientific research. 

This research is the ground for further work in the field of hate speech detection within youth language. The next step is to identify a non youth language online chat conversation and annotate it for hate speech, comparing the differences in language and use of hateful terms. Overall, the research can be used to train youth language specific hate speech classifiers and identify the influence of youth language on their performance. This research opens up the possibility to analyze the bias youth language introduces into existing pretrained hate speech detection models. Further, the generalizability of existing prediction models can be tested and increased by using this new data set.


\bibliographystyle{splncs04}
\bibliography{main}

\begin{thebibliography}{10}
\providecommand{\url}[1]{\texttt{#1}}
\providecommand{\urlprefix}{URL }
\providecommand{\doi}[1]{https://doi.org/#1}

\bibitem{Bartalesi2006}
Bartalesi, V., Sprugnoli, R., Lenzi, V.B., Moretti, G.: Cat: the celct
  annotation tool creep (cyberbullying effects prevention) view project
  it-timebank view project cat: the celct annotation tool (2006),
  \url{http://knowtator.sourceforge.net/index.shtml}

\bibitem{Bayzick2011}
Bayzick, J.: Detecting the presence of cyberbullying using computer software
  submitted to the faculty of ursinus college in fulfillment of the
  requirements for distinguished honors in computer science (2011)

\bibitem{Blodgett2017}
Blodgett, S.L., O'Connor, B.: Racial disparity in natural language processing:
  A case study of social media african-american english (2017).
  \doi{10.48550/ARXIV.1707.00061}, \url{https://arxiv.org/abs/1707.00061}

\bibitem{Chen2012}
Chen, Y., Zhou, Y., Zhu, S., Xu, H.: Detecting offensive language in social
  media to protect adolescent online safety. pp. 71--80 (2012).
  \doi{10.1109/SocialCom-PASSAT.2012.55}, 2012 ASE/IEEE International
  Conference on Social Computing, SocialCom 2012

\bibitem{Chung2019}
Chung, Y.L., Kuzmenko, E., Tekiroglu, S.S., Guerini, M.: Conan -- counter
  narratives through nichesourcing: a multilingual dataset of responses to
  fight online hate speech  (10 2019). \doi{10.18653/v1/P19-1271},
  \url{http://arxiv.org/abs/1910.03270 http://dx.doi.org/10.18653/v1/P19-1271}

\bibitem{Dadvar2013}
Dadvar, M., Trieschnigg, D., Ordelman, R., de~Jong, F.: Improving cyberbullying
  detection with user context. pp. pp 693--696 (1 2013).
  \doi{10.1007/978-3-642-36973-5_62}

\bibitem{Gebru2021}
Gebru, T., Morgenstern, J., Vecchione, B., Vaughan, J.W., Wallach, H., III,
  H.D., Crawford, K.: Datasheets for datasets. Commun. ACM  \textbf{64}(12),
  86–92 (nov 2021). \doi{10.1145/3458723},
  \url{https://doi.org/10.1145/3458723}

\bibitem{Gibert2018}
de~Gibert, O., Perez, N., García-Pablos, A., Cuadros, M.: Hate speech dataset
  from a white supremacy forum  (9 2018), \url{http://arxiv.org/abs/1809.04444}

\bibitem{gilbert2018}
de~Gibert, O., P{\'{e}}rez, N., Pablos, A.G., Cuadros, M.: Hate speech dataset
  from a white supremacy forum. CoRR  \textbf{abs/1809.04444} (2018),
  \url{http://arxiv.org/abs/1809.04444}

\bibitem{Hosseinmardi2015}
Hosseinmardi, H., Mattson, S.A., Ibn~Rafiq, R., Han, R., Lv, Q., Mishra, S.:
  Analyzing labeled cyberbullying incidents on the instagram social network.
  In: Liu, T.Y., Scollon, C.N., Zhu, W. (eds.) Social Informatics. pp. 49--66.
  Springer International Publishing, Cham (2015)

\bibitem{Hovy2021}
Hovy, D., Prabhumoye, S.: Five sources of bias in natural language processing.
  Language and Linguistics Compass  \textbf{15},  e12432 (2021).
  \doi{https://doi.org/10.1111/lnc3.12432},
  \url{https://compass.onlinelibrary.wiley.com/doi/abs/10.1111/lnc3.12432}

\bibitem{Kennedy2020}
Kennedy, B., Jin, X., Davani, A.M., Dehghani, M., Ren, X.: Contextualizing hate
  speech classifiers with post-hoc explanation. pp. 5435--5442. Association for
  Computational Linguistics (7 2020). \doi{10.18653/v1/2020.acl-main.483},
  \url{https://aclanthology.org/2020.acl-main.483}

\bibitem{Kurita2019}
Kurita, K., Vyas, N., Pareek, A., Black, A.W., Tsvetkov, Y.: Measuring bias in
  contextualized word representations. pp. 166--172. Association for
  Computational Linguistics (8 2019). \doi{10.18653/v1/W19-3823},
  \url{https://aclanthology.org/W19-3823}

\bibitem{Menini2019}
Menini, S., Moretti, G., Corazza, M., Cabrio, E., Tonelli, S., Villata, S.,
  Fondazione, Kessler, B.: A system to monitor cyberbullying based on message
  classification and social network analysis (2019),
  \url{https://fasttext.cc/docs/en/}

\bibitem{PaaschColberg2021}
Paasch-Colberg, S., Strippel, C., Trebbe, J., Emmer, M.: From insult to hate
  speech: Mapping offensive language in german user comments on immigration.
  Media and Communication  \textbf{9},  171--180 (02 2021).
  \doi{10.17645/mac.v9i1.3399}

\bibitem{Park2018}
Park, J.H., Shin, J., Fung, P.: Reducing gender bias in abusive language
  detection. pp. 2799--2804. Association for Computational Linguistics (10
  2018). \doi{10.18653/v1/D18-1302}, \url{https://aclanthology.org/D18-1302}

\bibitem{Saha2019}
Saha, K., Chandrasekharan, E., Choudhury, M.D.: Prevalence and psychological
  effects of hateful speech in online college communities (2019),
  \url{https://dl.acm.org/citation.cfm?id=3326032}

\bibitem{Schwartz2013}
Schwartz, H.A., Eichstaedt, J.C., Kern, M.L., Dziurzynski, L., Ramones, S.M.,
  Agrawal, M., Shah, A., Kosinski, M., Stillwell, D., Seligman, M.E., Ungar,
  L.H.: Personality, gender, and age in the language of social media: The
  open-vocabulary approach. PLoS ONE  \textbf{8} (9 2013).
  \doi{10.1371/journal.pone.0073791}

\bibitem{Sprugnoli2018}
Sprugnoli, R., Menini, S., Tonelli, S., Oncini, F., Piras, E.M., Kessler, F.B.:
  Creating a whatsapp dataset to study pre-teen cyberbullying (2018),
  \url{http://creep-project.eu/}

\bibitem{Thapa_Subedi_2018}
Thapa, R., Subedi, S.: Social media and depression. Journal of Psychiatrists’
  Association of Nepal  \textbf{7}(2),  1–4 (Dec 2018).
  \doi{10.3126/jpan.v7i2.24607},
  \url{https://www.nepjol.info/index.php/JPAN/article/view/24607}

\bibitem{bertie2021}
Vidgen, B., Derczynski, L.: Directions in abusive language training data, a
  systematic review: Garbage in, garbage out. PLOS ONE  \textbf{15}(12),  1--32
  (12 2021). \doi{10.1371/journal.pone.0243300},
  \url{https://doi.org/10.1371/journal.pone.0243300}

\bibitem{Wijesiriwardene2020}
Wijesiriwardene, T., Inan, H., Kursuncu, U., Gaur, M., Shalin, V.L.,
  Thirunarayan, K., Sheth, A., Arpinar, I.B.: Alone: A dataset for toxic
  behavior among adolescents on twitter  (8 2020),
  \url{http://arxiv.org/abs/2008.06465}

\end{thebibliography}


\end{document}